\documentclass[letterpaper, 10 pt, journal, twoside]{IEEEtran}

\usepackage{graphicx}
\usepackage{booktabs}
\usepackage{multirow}
\usepackage{stfloats}
\usepackage{subcaption}
\usepackage[font=small,labelfont=bf]{caption}
\usepackage{hyperref}

\begin{document}

\title{DefGraspSim: Physics-based simulation of grasp outcomes for 3D deformable objects}

\author{Isabella Huang$^{1, 2}$, Yashraj Narang$^{2}$, Clemens Eppner$^{2}$, Balakumar Sundaralingam$^{2}$, Miles Macklin$^{2}$, \\ Ruzena Bajcsy$^{1}$, Tucker Hermans$^{2, 3}$, Dieter Fox$^{2, 4}$
\thanks{Manuscript received: October, 22, 2021; Revised January, 14, 2021; Accepted February, 18, 2022.}
\thanks{This paper was recommended for publication by Editor Markus Vincze upon evaluation of the Associate Editor and Reviewers' comments.
This work was supported by NSF Award \#1545126} 
\thanks{$^{1}$Department of Electrical Engineering and Computer Sciences, University of California, Berkeley, USA {\tt\footnotesize isabella.huang@berkeley.edu};$^{2}$NVIDIA Corporation, Seattle, USA;$^{3}$School of Computing, University of Utah, Salt Lake City, USA;$^{4}$Paul G. Allen School of Computer Science \& Engineering, University of Washington, Seattle, USA}%
\thanks{Digital Object Identifier 10.1109/LRA.2022.3158725}
}

\markboth{IEEE Robotics and Automation Letters. Preprint Version. Accepted March, 2022}
{Huang \MakeLowercase{\textit{et al.}}: DefGraspSim}

\maketitle

\begin{abstract}
Robotic grasping of 3D deformable objects (e.g., fruits/vegetables, internal organs, bottles/boxes) is critical for real-world applications such as food processing, robotic surgery, and household automation. However, developing grasp strategies for such objects is uniquely challenging. Unlike rigid objects, deformable objects have infinite degrees of freedom and require field quantities (e.g., deformation, stress) to fully define their state. As these quantities are not easily accessible in the real world, we propose studying interaction with deformable objects through physics-based simulation. As such, we simulate grasps on a wide range of 3D deformable objects using a GPU-based implementation of the corotational finite element method (FEM). To facilitate future research, we open-source our simulated dataset (34 objects, 1e5 Pa elasticity range, 6800 grasp evaluations, 1.1M grasp measurements), as well as a code repository that allows researchers to run our full FEM-based grasp evaluation pipeline on arbitrary 3D object models of their choice. Finally, we demonstrate good correspondence between grasp outcomes on simulated objects and their real counterparts.
\end{abstract}

\begin{IEEEkeywords}
Grasping, simulation and animation, software tools for benchmarking and reproducibility.
\end{IEEEkeywords}

\IEEEpeerreviewmaketitle

\section{Introduction}
\IEEEPARstart{F}{rom} clothing, to plastic bottles, to humans, deformable objects are omnipresent in our world. A large subset of these are \textit{3D deformable objects} (e.g., fruits, internal organs, and flexible containers), for which dimensions along all 3 spatial axes are of similar magnitude, and significant deformations can occur along any of them. Robotic grasping of 3D deformables is underexplored relative to rope and cloth, but remains critical for applications like food handling \cite{Gemici2014IROS}, robotic surgery \cite{Smolen2009ICACHI}, and domestic tasks \cite{Sanchez2018IJRR}. Compared to rigid objects, grasping 3D deformable objects faces 4 major challenges to which we respond with 4 key contributions.

First, classical analytical metrics for grasping rigid objects (e.g., force/form closure) do not typically consider deformation of the object during the grasp \cite{Sanchez2018IJRR}. Yet, deformations significantly impact the contact surface and object dynamics. For example, one can grasp a soft toy haphazardly; however, if the toy were rigid, it would no longer conform to one's hands, and many grasps would become unstable. Conversely, one can grasp a rigid container haphazardly; however, if the container were flexible, grasps along its faces would crush its contents. Unlike for rigid objects, the success of a 3D deformable grasp depends on properties such as compliance. We thus propose a set of diverse \textit{performance metrics} that quantify deformable grasp outcomes (Sec.~\ref{sec:metrics}), such as stability, deformation, and stress. Performance metrics may also compete (e.g., a stable grasp may induce high deformation).

\begin{figure}
\centering
\includegraphics[width=\linewidth, trim={0cm 0cm 0cm 0cm},clip]{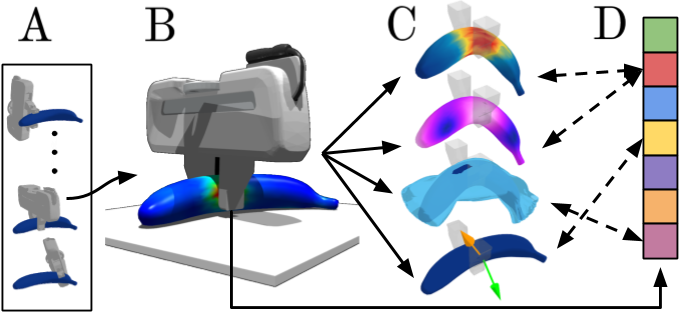}
\caption{(A) For a broad set of candidate grasps on a deformable object, (B) we simulate the object's response with FEM, (C) measure performance metrics (e.g., stress, deformation, controllability, instability), and (D) identify pre-pickup grasp features that are correlated with the metrics. Our simulated dataset contains 34 objects, 6800 grasp experiments, and $1.1M$ unique measurements. 
}
\label{fig:front_figure}
\vspace{-10pt}

\end{figure}
Second, performance metrics may be partially or fully unobservable (e.g., volumetric stress fields), requiring estimators in the real world. Previous works have addressed this by formulating quality metrics, which we refer to as \textit{grasp features}: simple quantities a robot can measure before pickup that can predict performance metrics. Whereas grasp features have predominantly been designed for rigid objects, we propose a set of grasp features compatible with deformables (Sec.~\ref{sec:features}). 


Third, there exists neither a general framework to evaluate arbitrary deformable grasps (i.e., via performance metrics and grasp features), nor an exhaustive dataset of deformable grasp experiments. We thus release DefGraspSim, a codebase that allows users to automatically perform an exhaustive set of FEM-based grasp evaluations on arbitrary 3D objects.\footnote{\url{https://sites.google.com/nvidia.com/defgraspsim}} Simulation offers multiple advantages: it extends classical analytical methods through accurate modeling of object deformation, enables safe execution of experiments, and provides full observability of performance metrics. We also conduct a large-scale simulation-based study of 3D deformable object grasping on 3D deformables varying in geometry and elasticity (Fig.~\ref{fig:front_figure}), and provide this live dataset of 34 objects, 6800 grasp evaluations, and $1.1M$ corresponding measurements. This is the largest deformable object grasping dataset in existence.

Fourth, simulation studies do not necessarily correspond to real-world behavior. To address this gap, we perform a pilot sim-to-real study on results generated by DefGraspSim, and demonstrate that simulated results show reliable correspondence with real-world experiments (Sec.~\ref{sec:sim2real}).

We believe these 4 main contributions are an important milestone towards developing a complete learning and planning framework for grasping 3D deformables.

\begin{table*}
  \centering
  \caption{Comparisons between Isaac Gym and other robotics simulators that support both 3D deformable bodies and actuator interactions.}

  \begin{tabular}{p{2.2cm}p{1.45cm}p{2.0cm}p{1.9cm}p{2.5cm}p{4.0cm}p{1.0cm}} \toprule
    \textbf{Simulator} & \textbf{Interactions} & \textbf{3D geometries} & \textbf{Materials} & \textbf{Underlying model} & \textbf{Observable states} & \textbf{Processor} \\
    \midrule 
    MuJoCo~\cite{Todorov2012IROS}
    & soft-rigid, rigid-rigid
    & Boxes, cylinders, ellipsoids
    & Homogeneous isotropic elastic 
    & Mass-spring with surface nodes
    & Nodal positions 
    & CPU \\
    PyBullet 3~\cite{coumans2019}
    & soft-rigid, soft-soft, rigid-rigid
    & Arbitrary geometries
    & Homogeneous isotropic elastic/hyperelastic 
    & Mass-spring or Neo-Hookean volumetric FEM
    & Nodal positions, contact points \& forces
    & CPU \\
    IPC-GraspSim~\cite{kim2021IPC}
    & soft-soft
    & Arbitrary geometries
    & Homogeneous isotropic elastic 
    & Incremental potential contact model
    & Nodal positions, velocities, and accelerations
    & CPU \\
    Isaac Gym~\cite{makoviychuk2021isaac}
    & soft-rigid, rigid-rigid 
    & Arbitrary geometries
    & Homogeneous isotropic elastic 
    & Co-rotational linear volumetric FEM
    & Nodal positions \& velocities, contact points \& forces, element stress tensors
    & GPU \\
    \bottomrule 

  \end{tabular}
  \label{tab:sim_compare}
\end{table*}

\section{Related Work}\label{sec:related_works}
\noindent\textbf{Modeling techniques}. With over three decades of development, methods in rigid-object grasp planning range from model-based approaches using exact geometries \cite{Sastry1988IJRA,Ferrari1992ICRA,Miller2003ICRA} to data-driven approaches without full models \cite{Lenz2013IJRR,Dang2012a,Mahler2017CORR,Kopicki2016,mousavian2019,lu-ram2020-grasp-inference}. Rigid-body grasping simulators such as GraspIt!~\cite{Miller2004RAM} and OpenGRASP~\cite{Len2010OpenGRASP} have been used to develop many of these algorithms. 
For 3D deformable objects, rigid-body approximations can lead to efficient simulations \cite{pozziefficient}; however, continuum models are preferred, as they can represent large deformations and allow consistent material parameters without an explicit model-fitting stage \cite{duriez.13}. 3D continuum models include Kelvin-Voigt elements governed by nonlinear PDEs~\cite{Howard2000AR}, mass-spring models~\cite{Lazher2014ICR}, 2D FEM for planar and ring-like objects~\cite{Jia2014IJRR}, and gold-standard 3D FEM~\cite{Lin2015IJRR}. However, many powerful FEM simulators used in engineering and graphics (e.g., Vega~\cite{Vega}) do not feature infrastructure for robotic control, such as built-in joint control. For comprehensive reviews of 3D deformable modeling techniques, please refer to \cite{Arriola2020FRAI, YinScienceRobotics2021}. In this work, we use the GPU-accelerated Isaac Gym simulator to analyze grasp interactions with deformable objects. In Table~\ref{tab:sim_compare}, we compare Isaac Gym to other robotics simulators, including MuJoCo~\cite{Todorov2012IROS} and PyBullet~\cite{coumans2019}, which have successfully modeled deformable ropes and cloths using rigid-body networks with compliant constraints \cite{Wang2015ICRA,Maitin-Shepard2010ICRA,Li2015IROS,Clegg2020RAL}, but have recently started to support 3D deformable objects as well.

\noindent\textbf{Performance metrics}. Prior works have evaluated 3D deformable-object grasps using performance metrics based on pickup success, strain energy, deformation, and stress. 
Success-based metrics include the minimum force required by a particular grasp, which is calculated via real-world iterative search~\cite{Howard2000AR} and FEM~\cite{Lazher2014ICR,Lin2015IJRR}. Success depends on both object geometry and stiffness (e.g., a cone can be picked up only when it can deform to the gripper) \cite{Lin2015IJRR}. Metrics based on strain energy (i.e., elastic potential energy in the object) have served as proxies for an object's stability against external wrenches. In 2D the deform closure metric generalizes rigid form closure \cite{Bicchi1995IJRR} and quantifies the positive work required to release an object from a grasp \cite{Goldberg2005IJRR}. It is optimized by maximizing strain energy without inducing plastic deformation. Similarly, for thin and planar 2.5D objects, grasps have been selected to maximize strain energy under a fixed squeezing distance~\cite{Jia2014IJRR}. Deformation-based metrics have also been proposed for cups and bottles to detect whether contents are dislodged during lifting and rotation \cite{Xu2020ICRA}. Finally, stress-based metrics have been proposed to avoid material fracture, but were evaluated only on rigid objects \cite{Pan2020ICRA}. 

\noindent\textbf{Grasp features}. Many grasp features to predict grasp performance have been previously investigated on rigid objects. Features include force and form closure \cite{Ferrari1992ICRA} and grasp polygon area \cite{Mirtich1994ICRA}, and their predictive accuracy has been tested under different classification models \cite{Rubert2017IROS}. A thorough survey on rigid grasping features can be found in \cite{Roa2014AR}. However, grasp features for deformables have only been explored in one study, which measured the work performed on containers during grasping to predict whether its liquid contents would be displaced \cite{Xu2020ICRA}.

\begin{figure}
\centering
\frame{\includegraphics[width=\linewidth]{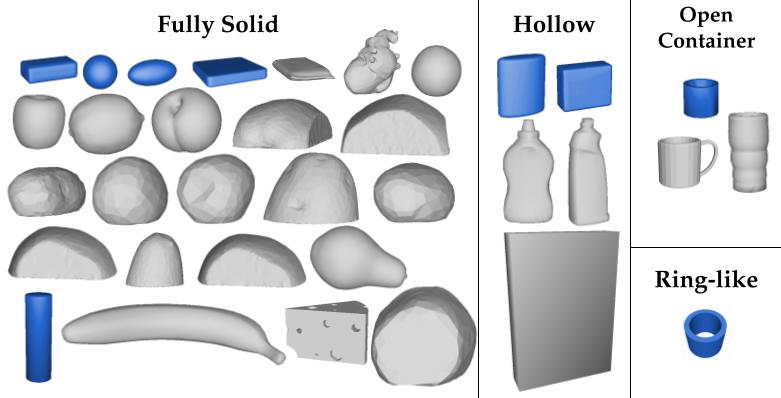}}
\caption{The 34 evaluated objects grouped by geometry and dimension (shown to scale). Objects in blue are self-designed primitives; those in gray are scaled models from open datasets \cite{Calli2015RAM,Wu2015CVPR,thingiverse, ybjDataset}.
}
\label{fig:object_categories}
\vspace{-10pt}
\end{figure}

\begin{figure*}
\centering
\includegraphics[width=\textwidth]{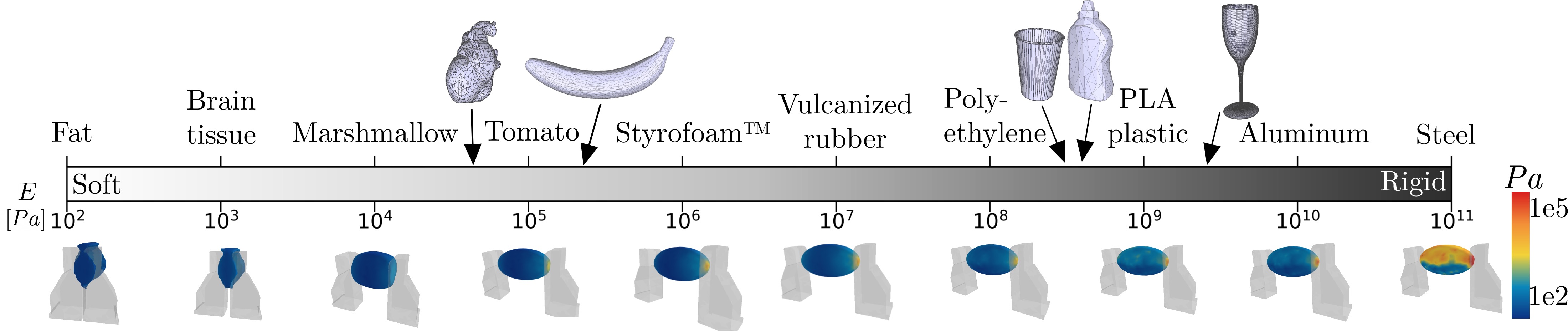}
\caption{Young's modulus $E$ for various materials (adapted from \cite{soro2013SR}). (Top): real-world objects and their typical $E$. (Bottom): Stress distributions of an ellipsoid under $1$~$N$ of grasp force. Soft ellipsoids undergo large deformations; rigid ones have high-stress regions. 
}
\label{fig:spectrum}
\vspace{-5pt}
\end{figure*}

\section{Grasp Simulator} \label{sec:sim}
We use the FEM-based simulator Isaac Gym \cite{makoviychuk2021isaac} to simulate grasps on 3D deformable objects. FEM is a variational numerical technique that divides complex geometrical domains into simple subregions and solves the weak form of the governing partial differential equations over each region.
In FEM simulation, a deformable object is represented by a volumetric mesh of \textit{elements}; the object's configuration is described by the element vertices, known as \textit{nodes}. We use Isaac Gym's \cite{makoviychuk2021isaac} co-rotational linear constitutive model of the object's internal dynamics coupled to a rigid-body representation of the robotic gripper via an isotropic Coulomb contact model~\cite{stewart2000rigid}. A GPU-based Newton method performs implicit integration by solving a nonlinear complementarity problem~\cite{macklin2019}. 
At each timestep, the simulator returns element stress tensors and nodal positions, which are used to calculate grasp metrics. With sufficiently small timesteps and high mesh density, FEM predictions for deformable solids can be extremely accurate \cite{Reddy2019Book, Narang2020RSS}. We simulate at a frequency of $1500 Hz$. The simulator executes at $5$-$10$~$fps$, and each grasp experiment (Sec.~\ref{sec:grasp_experiments}) requires around 2 to 7 minutes to run. 

We evaluate a set of 34 3D deformable objects comprising both simple object primitives and complex real-world models, categorized by geometry and dimension (Fig. \ref{fig:object_categories}). We process object surface meshes in Blender to smooth sharp edges to avoid stress singularities, and reduce node count where possible to optimize speed. We then convert these into tetrahedral meshes using fTetWild \cite{ftw}. 

For all experiments our objects have density $\rho = 1000 \frac{kg}{m^3}$, Poisson's ratio $\nu = 0.3$, coefficient of friction $\mu = 0.7$, and Young's modulus $E \in \mathcal{E} = \{2e4, 2e5, 2e6, 2e9\} Pa$. $\mathcal{E}$ covers a wide range of real materials, from human skin ($\sim$$10^4 Pa$) to ABS plastic ($\sim$$10^9 Pa$) (Fig.~\ref{fig:spectrum}). 
The target squeezing force on an object is $F_{p} = 1.3 \times \frac{mg}{\mu}$ (where $m$ is mass and $g$ is gravity), which is the force required to support the object's weight with a factor of safety. For a fixed $E$, increasing $\mu$ decreases $F_p$ as well as the induced deformation. This effect is essentially the same as if $\mu$ is fixed while $E$ is increased, since an elastically stiffer object will also deform less for the same $F_p$ applied. Thus, we fix $\mu$ and vary $E$.

\section{Grasp Experiments} \label{sec:grasp_experiments}

\begin{figure}
\centering
\includegraphics[scale=0.4]{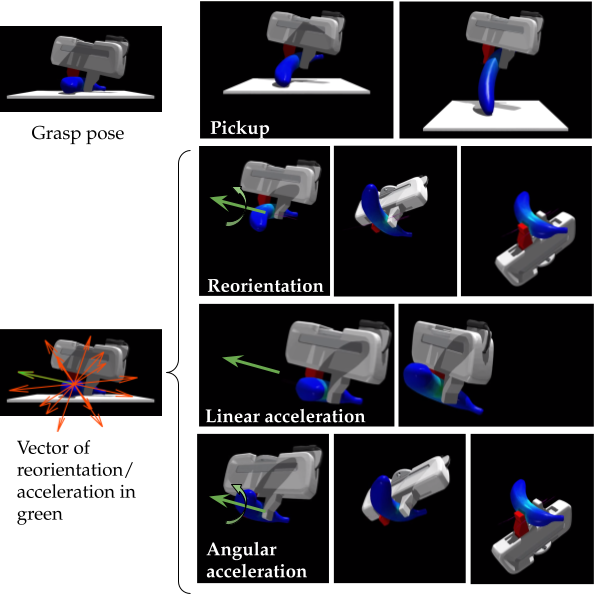}
\caption{Example frames from the execution of four different experiments per grasp on a banana: pickup, reorient, twist (angular acceleration), and shake (linear acceleration).} 
\label{fig:grasp_tests}
\end{figure}

We perform simulated grasping experiments within Isaac Gym on 34 objects using the widely used Franka parallel-jaw gripper. To generalize to other parallel-jaw grippers we remove all specialized gripper features.
For each object, we generate a diverse set of 50 candidate grasps with an antipodal sampler~\cite{EppnerISRR2019}. Each object initially rests atop a horizontal plane; we disable gripper collisions with the plane to test the full spatial distribution of grasps by allowing grasps to come from underneath. Prior to grasping, the pre-contact nodal positions and element stresses of the object are recorded. The gripper is initialized at a candidate grasp pose, then squeezes using a force-based torque controller to achieve the target grasp force, $F_p$. 
Once $F_p$ converges, the grasp features are measured. Then, one of the following experiments (Fig.~\ref{fig:grasp_tests}) is executed: \textit{pickup, reorientation, linear acceleration, and angular acceleration}.

\noindent\textbf{1) Pickup}. The platform lowers to apply incremental gravitational loading to the object. Pickup is a success if the gripper maintains contact with the object for 5 seconds. If so, stress and deformation fields are recorded, and stress, deformation, and strain energy performance metrics are computed.

\noindent\textbf{2) Reorientation}. The grasp force is increased from $F_{p}$ to $F_{slip}$, the minimum force required to counteract rotational slip. The platform is lowered until the object is picked up. The gripper rotates the object to 64 unique reorientation states. We record stress and deformation fields at each state, and compute deformation controllability as the maximum deformation over all states. $F_{slip}$ is estimated by approximating each gripper contact patch as 2 point-contacts that oppose the gravitational moment. The gripper rotates the object about each of 16 vectors regularly spaced in a unit 2-sphere at angles $k\pi/4, k \in [1..4]$ for a total of 64 unique reorientation states.

\noindent\textbf{3) Linear acceleration}.
The gripper linearly accelerates along each of 16 unique direction vectors as in the reorientation experiment. Each vector has a complement pointing in the opposite direction; thus, this method generalizes the cyclic shaking tests from previous works~\cite{EppnerISRR2019}. The acceleration is recorded at which at least one finger loses contact with the object. Linear instability is computed as the average loss-of-contact acceleration over all directions. The robot moves at $1000 \frac{m}{s}^3$ jerk in a gravity-free environment, corresponding to a linearly increasing acceleration. We impose a realistic upper acceleration limit of $50 \frac{m}{s}^2$ ($\approx 5g$). 

\noindent\textbf{4) Angular acceleration}. The gripper rotationally accelerates about 16 unique axes. Angular instability is computed as the average loss-of-contact acceleration over all axes. The robot accelerates at $2500 \frac{rad}{s}^3$ jerk; to mitigate undesired linear acceleration, the midpoint between the fingers is the center of rotation. The angular loss-of-contact threshold is limited to $1000 \frac{rad}{s}^2$ (i.e., the linear acceleration limit, scaled by the $0.04 m$ max. finger displacement, a reference moment arm).

\noindent\textbf{Controller Details.} A contact force-based torque controller is used to achieve the desired grasp forces. A low-pass filter is first applied to the contact force signals due to high frequency noise that prevails from small numerical fluctuations in position, especially at higher moduli. For the three experiments involving post-pickup manipulation, the finger joints are frozen immediately after pickup to maintain the gripper separation.

\begin{figure}
\centering
\includegraphics[width=\linewidth, trim={0cm 0cm 0cm 0cm},clip]{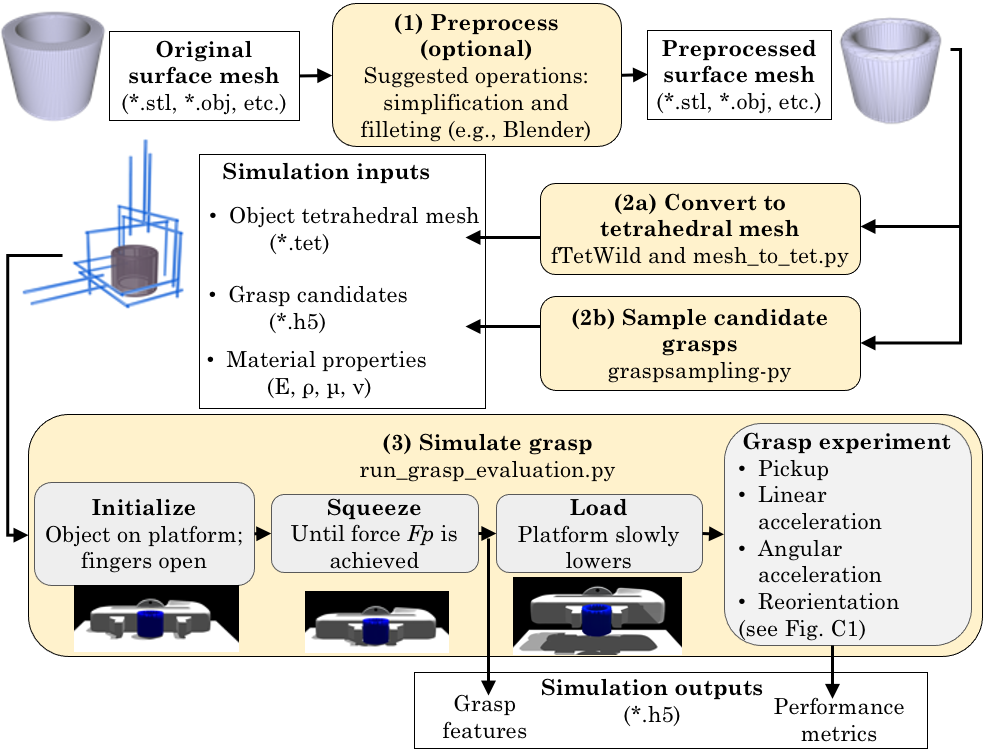}
\caption{Software flow diagram of the DefGraspSim codebase.
}
\label{fig:software_flow}
\end{figure}

\noindent\textbf{Codebase}. We release the code to replicate our experiments on arbitrary objects,\footnote{\url{https://github.com/NVlabs/DefGraspSim}} and a full software flow diagram is shown in Fig.~\ref{fig:software_flow}. Candidate grasps can either be user-defined or generated using our grasp sampling module. We also encourage the use of external software such as Blender to preprocess object models (e.g., mesh simplification and edge filleting) to improve simulation speed and convergence.

\section{Grasp Performance Metrics}

\label{sec:metrics}

During the preceding experiments, we measure the following 7 performance metrics to comprehensively evaluate grasp outcomes. These metrics include high-dimensional field quantities (e.g., stress fields), which are unexplored in rigid object modeling. Note that linear and angular instability are separate metrics. During each grasp experiment, we also measure a set of grasp features (Sec.~\ref{sec:features}), which capture low-dimensional state information about the grasp and have potential to be used as predictors of the final performance metrics. 

\noindent\textbf{Pickup success:} A binary metric measuring whether an object is lifted from a support plane. 

\noindent\textbf{Stress:} The element-wise stress field of an object when picked up. Exceeding material thresholds (e.g., yield stress, ultimate stress) leads to permanent deformation, damage, or fracture; examples include creasing of boxes, bruising of fruit, 
and perforation of organs. We convert each element's stress tensor into von Mises stress, a scalar quantity that quantifies whether an element has exceeded its yield threshold. We then measure the maximum stress over all elements, since real-world applications typically aim to avoid damage at any point.

\noindent\textbf{Deformation:} The node-wise displacement field of the object from pre- to post-pickup, neglecting rigid-body transformations. Deformation must often be minimized (e.g., on flexible containers with contents that can be damaged or dislodged). To compute this field, the difference between the pre- and post-pickup nodal positions is calculated, the closest rigid transform is determined \cite{hornung2017}, and the transform is subtracted. We compute the $\ell^2$ norm of each node's displacement and measure the maximum value over all nodes.

\noindent\textbf{Strain energy:} The elastic potential energy stored in the object (analogous to a Hookean spring). Conveniently, this metric penalizes both stress and deformation. The strain energy is given by $U_e = \int_V \sigma^T \epsilon dV$, where $\sigma$, $\epsilon$, and $V$ are the stress tensor, strain tensor, and volume, respectively.

\noindent\textbf{Linear and angular instability:} We define instability as the minimum acceleration applied to the gripper (\textit{along} or \textit{about} a vector for linear and angular instability, respectively) at which the object loses contact (i.e., separates along the gripper normal, or slides out of the gripper). This measures how easily an object is displaced from the grasp under external forces.

\begin{figure}
\centering
\includegraphics[scale=0.11]{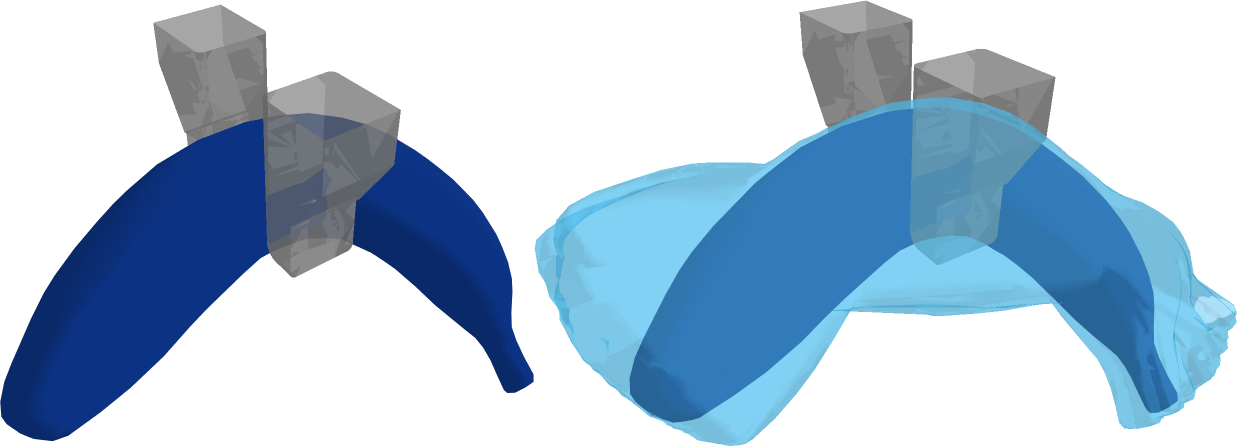}
\caption{Illustration of deformation controllability. A soft banana-shaped object under pickup (left); the union of all shape configurations achieved under reorientation, superimposed in light blue (right). 
}
\label{fig:def_banana}
\end{figure}

\noindent\textbf{Deformation controllability:} We define deformation controllability as the maximum deformation when the object is reoriented under gravity. (An example of shape change induced during reorientation is shown in Fig.~\ref{fig:def_banana}.) Depending on the task, it may be useful to either minimize or maximize deformation controllability. For example, to reduce the effects of post-grasp reorientation on deformation, minimizing this metric allows the object to behave rigidly after pickup. Alternatively, to augment the effects of post-grasp reorientation (e.g., during insertion of endoscopes), we may maximize it instead. Our notion of deformation controllability is different from the classical notion (i.e., the ability to achieve any robot state in finite time). Here, we are not modifying robot controllability by changing actuation, but modifying object controllability by changing the number of possible deformation states. 

\section{Grasp Features}\label{sec:features}

\begin{figure}
     \centering
     \begin{subfigure}[b]{.49\columnwidth}
         \centering
         \includegraphics[scale=0.21, trim={0cm 4cm 15cm 5.5cm},clip]{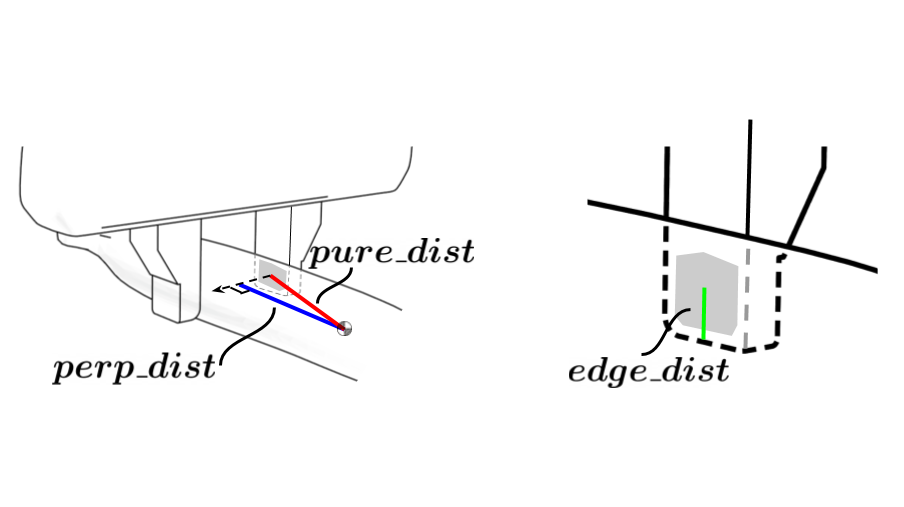}
         \caption{}
         \label{fig:qm1}
     \end{subfigure}
     \hfill
     \begin{subfigure}[b]{.49\columnwidth}
         \centering
         \includegraphics[scale=0.21, trim={20cm 4cm 3.5cm 5cm},clip]{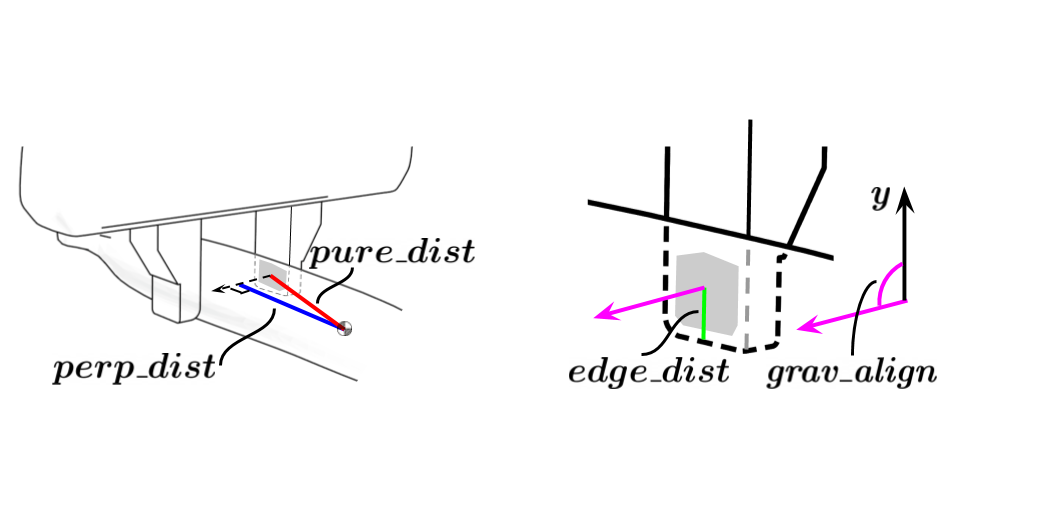}
         \caption{}
         \label{fig:qm2}
     \end{subfigure}
     \hfill
        \caption{Four grasp features illustrated on a Franka gripper.
        }
        \label{fig:qm}
\end{figure}

The 7 grasp features are recorded after applying the grasping force $F_p$, but before loading (Fig.~\ref{fig:software_flow}). All can be measured by common real-world sensors (e.g., encoders, cameras, and tactile arrays) and are summarized in Table~\ref{tab:1} and Fig.~\ref{fig:qm}, along with references to existing works from which they are derived. See \cite{Roa2014AR} for a full review of grasp features on rigid objects.

\begin{table*}
  \centering
  \caption{Grasp features, their descriptions, and existing works from which they are derived.}

  \begin{tabular}{p{4cm}p{1.5cm}p{8.6cm}p{2.4cm}}
    \toprule
    \textbf{Feature} & \textbf{Abbreviation} & \textbf{Definition and Relevance} & \textbf{Usage in Literature} \\
    \midrule 
    Contact patch distance to centroid 
    & \textit{pure\_dist} 
    & Distance from the center of each finger's contact patch to the object's center of mass (COM) (Fig. \ref{fig:qm}\subref{fig:qm1}), averaged over the two fingers.
    & \cite{Rubert2017IROS, DingITRA2001} \\
    Contact patch perpendicular distance to centroid 
    & \textit{perp\_dist} 
    & Perpendicular distance from the center of each finger's contact patch to the object's COM (Fig. \ref{fig:qm}\subref{fig:qm1}), averaged over the two fingers; quantifies distance from lines of action of squeezing force.
    & \cite{BalasubramanianICRA2010} \\
    Number of contact points 
    & \textit{num\_contacts} 
    & Number of contact points on each finger, averaged over the fingers; quantifies amount of contact made.
    & \cite{Rubert2017IROS, DingITRA2001} \\
    Contact patch distance to finger edge 
    & \textit{edge\_dist} 
    & Distance from each finger's distal edge to the center of its contact patch (Fig. \ref{fig:qm}\subref{fig:qm2}), averaged over the two fingers.
    & \cite{Feix2016ITHMS} \\
    Gripper squeezing distance
    & \textit{squeeze\_dist} 
    & Change in finger separation from initial contact to the point at which $F_p$ is achieved; quantifies local deformation applied to the object.
    & \cite{Xu2020ICRA} \\
    Gripper separation 
    & \textit{gripper\_sep} 
    & Finger separation upon achieving $F_p$; quantifies the thickness of material between the fingers at grasp.
    & \cite{Rubert2017IROS} \\
    Alignment with gravity 
    & \textit{grav\_align} 
    & Angle between the finger normal and the global vertical; grounds the grasp pose to a fixed frame (Fig. \ref{fig:qm}\subref{fig:qm2}).
    & \cite{Vina2016ICRA} \\
    \bottomrule
  \end{tabular}
  \label{tab:1}
\end{table*}

\section{Example Simulation Results}\label{sec:software}
\begin{figure*}
     \centering
     \begin{subfigure}[b]{.27\textwidth}
         \centering
        \includegraphics[scale=0.12, trim={0cm 0cm 111cm 0cm},clip]{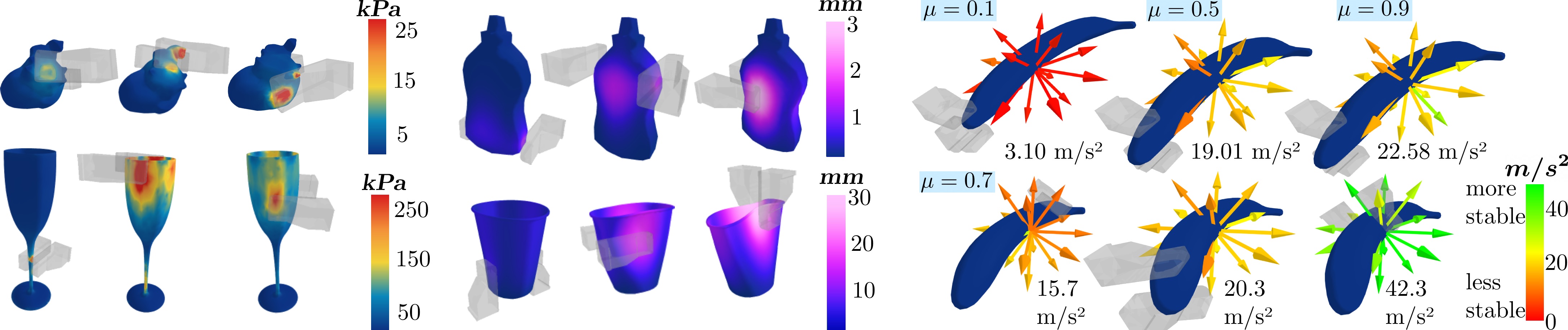}
        \caption{Simulated stress fields after pickup for various grasps on a (top) heart and (bottom) wine glass. Objects are colored by the von Mises stress field.
        \newline
        \newline
        }
         \label{fig:stress_examples}
     \end{subfigure}
     \hfill
     \begin{subfigure}[b]{.27\textwidth}
         \centering
        \includegraphics[scale=0.12, trim={45cm 0cm 67.5cm 0cm},clip]{figures/stress_def.jpg}
        \caption{Simulated deformation fields after pickup for various grasps on a (top) mustard bottle and (bottom) plastic cup. Objects are colored by the $l^2$-norm of the deformation field.
        \newline
        }
         \label{fig:def_examples}
     \end{subfigure}
     \hfill
     \begin{subfigure}[b]{.42\textwidth}
         \centering
        \includegraphics[scale=0.12, trim={90cm 0cm 0cm 0cm},clip]{figures/stress_def.jpg}
        \caption{Linear stability of grasps on a banana at $4$~$N$ of grasp force under (top) the same grasp but variable friction $\mu$, and (bottom) the same $\mu$ but variable grasps. Arrows are colored by the maximum acceleration in that direction before loss of contact. Number indicates the average acceleration at failure over all 16 directions.
        }
         \label{fig:stability_examples}
     \end{subfigure}
        \caption{Examples of simulated grasp outcomes on 5 objects, with visualizations of (a) stress, (b) deformation, and (c) linear stability.
        }
        \label{fig:full_examples}
\end{figure*}

To demonstrate the utility of DefGraspSim, we simulate grasps on 5 real object models with realistic material parameters (Fig.~\ref{fig:spectrum}, top) and visualize the resulting performance metrics of stress, deformation, and linear stability (Fig.~\ref{fig:full_examples}). The results are well-aligned with established mechanics principles. On the heart and wine glass (Fig.~\ref{fig:stress_examples}), regions of high stress arise with reduced contact areas (e.g., at heart nodules and the stem of the glass) and curvature discontinuities (e.g., at the lip of the glass). On a mustard bottle and plastic cup (Fig.~\ref{fig:def_examples}), high deformations occur when contacting regions of low geometric stiffness (e.g., at the main face of the bottle and lip of the cup) and vice versa (e.g. at the base of the bottle and cup). On a banana (Fig.~\ref{fig:stability_examples}), stability increases with friction $\mu$ under the same grasp pose and force. When $\mu$ is fixed, grasps closest to the ends of the fruit are least stable.

\begin{figure}
\centering
\includegraphics[width=0.96\linewidth, trim={0cm 0cm 0cm 0cm},clip]{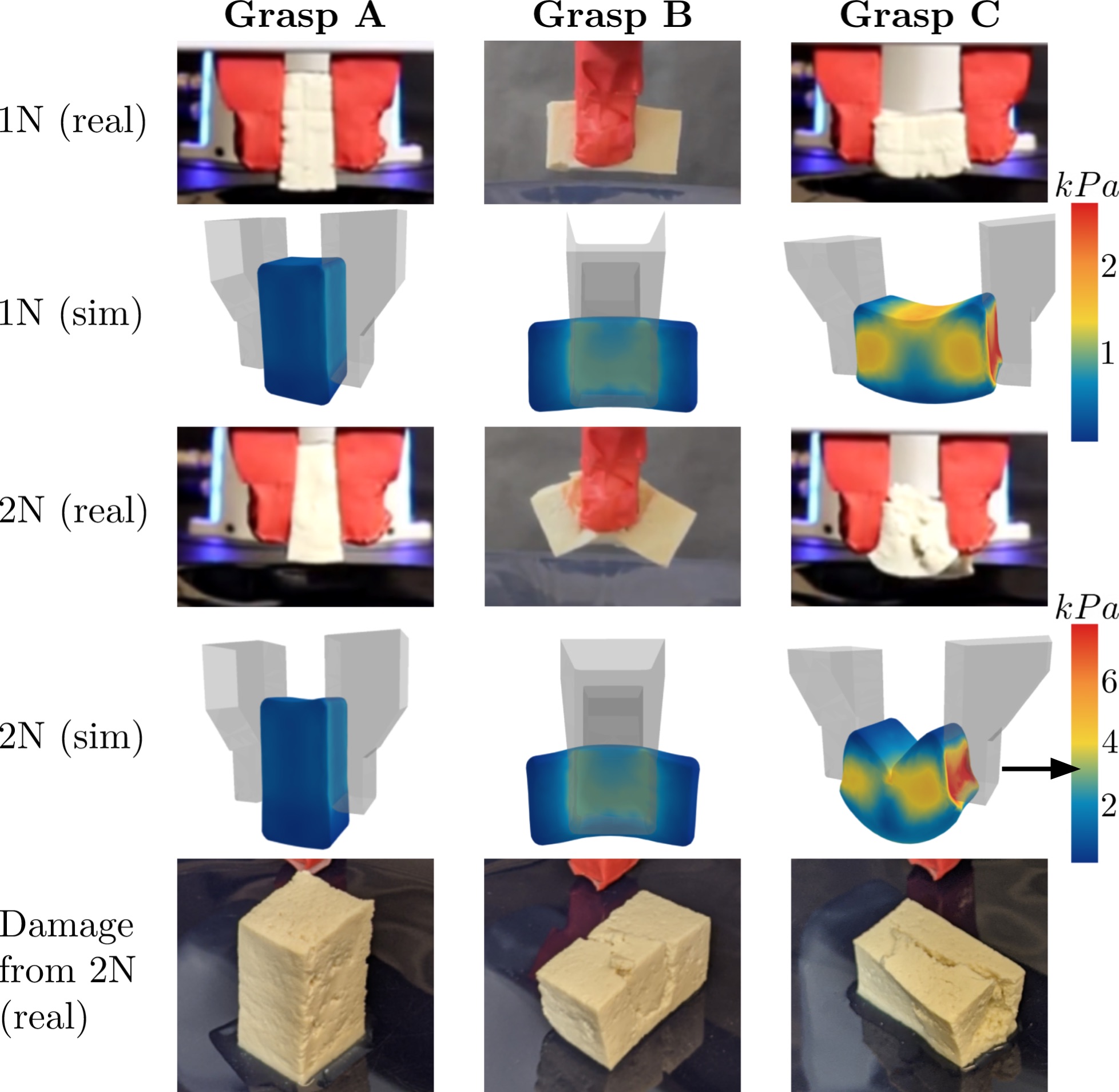}
\caption{Three grasps tested on blocks of tofu (1 and 2~$N$ of squeezing force) show similar outcomes in simulation and the real world. Real areas of fracture correspond to simulated stress greater than 3~$kPa$, the estimated breaking stress (denoted on color bar by black arrow).}
\label{fig:tofu_grasps}
\end{figure}

\section{Sim-to-Real Accuracy}\label{sec:sim2real}

Although simulation is the primary focus of this work, we also investigate whether grasp outcomes predicted by DefGraspSim are faithful to the real world. 

\noindent\textbf{Tofu blocks}. First, we test 3 grasps on real blocks of tofu under $1$ and $2$~$N$ of applied force~(Fig.~\ref{fig:tofu_grasps}). Simulated and real-world deformations exhibit strong similarities, and grasps achieve anticipated performance (e.g., grasps~A and~C,  respectively, minimize and maximize sagging under the 2 force conditions). Also, permanent damage on the real-world tofu occurs under 2~N of applied force, with fracture occurring in grasps~B and~C. Although standard FEM cannot simulate fracture, simulated stresses for these grasps lie within the literature-reported range of breaking stress for tofu, around 3$~kPa$~\cite{Toda2003SeedPC}. Moreover, fracture lines on the real tofu coincide with regions of stress higher than this threshold (i.e., along gripper edge under Grasp B; at tofu ends under Grasp C).

\begin{figure}
\centering
\includegraphics[width=\linewidth, trim={0cm 0cm 0cm 0cm},clip]{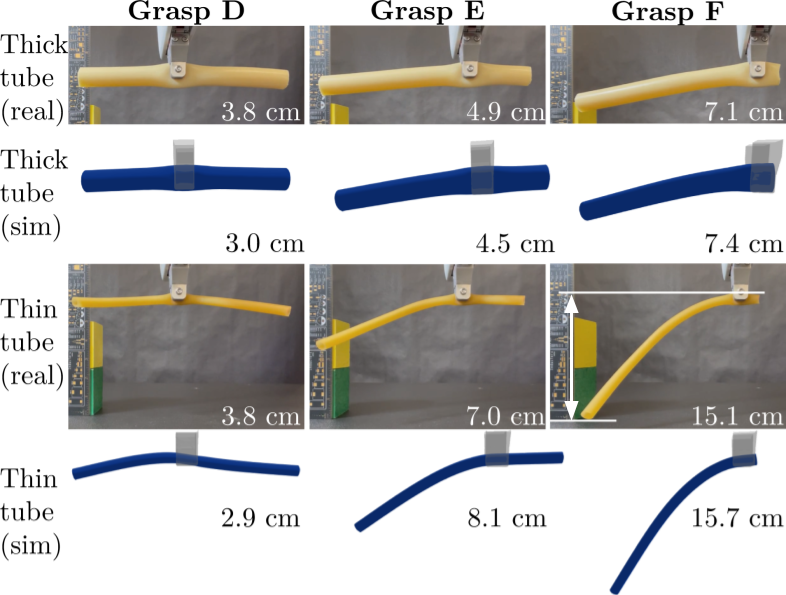}
\caption{Three grasps tested on 2 real and simulated latex tubes under 15 N of gripper force. The vertical distance between the highest and lowest points of the tube is annotated. Localized deformation due to compression at the grippers is replicated in simulation.}
\label{fig:hollow_tube_grasps}
\end{figure}

\begin{figure}
\centering
         \includegraphics[scale=0.3, trim={0cm 0cm 0cm 0cm},clip]{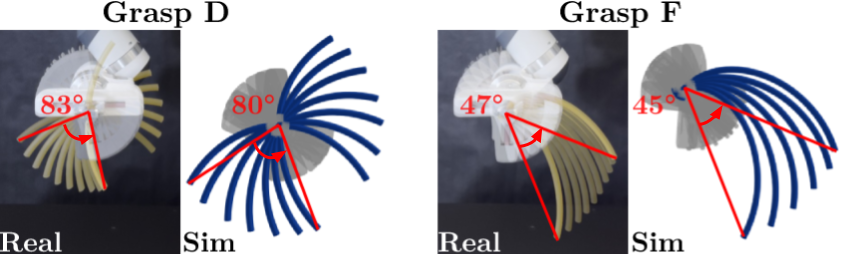}
         \caption{A middle grasp (grasp D) and end grasp (grasp F) under a counterclockwise 90$^\circ$ rotation of the gripper in the real world and in simulation. The angles swept out by the tube tip are marked in red.
}
\label{fig:tube_reorient}
\end{figure}

\noindent\textbf{Latex tubes}. Next, we perform 3 grasps on latex tubes of different geometry (Fig.~\ref{fig:hollow_tube_grasps}). Again, simulated and real-world deformations are highly similar, including indentations and bulges localized to regions of contact; moreover, the vertical distance between the highest and lowest points of the tubes closely match. We also test the deformation controllability between grasp D (a middle grasp) and grasp F (an end grasp) on the thin tube (Fig.~\ref{fig:tube_reorient}) by rotating the gripper by 90 degrees under each grasp. Deformation controllability is higher in grasp F, as the resulting angle swept out by the tube tip is only 47$^\circ$ (compared to 83$^\circ$ under grasp D), which indicates that more shape change is induced. These angle values also closely match those predicted by simulation.

\begin{figure}
\centering
\includegraphics[width=\linewidth, trim={0cm 0cm 0cm 0cm},clip]{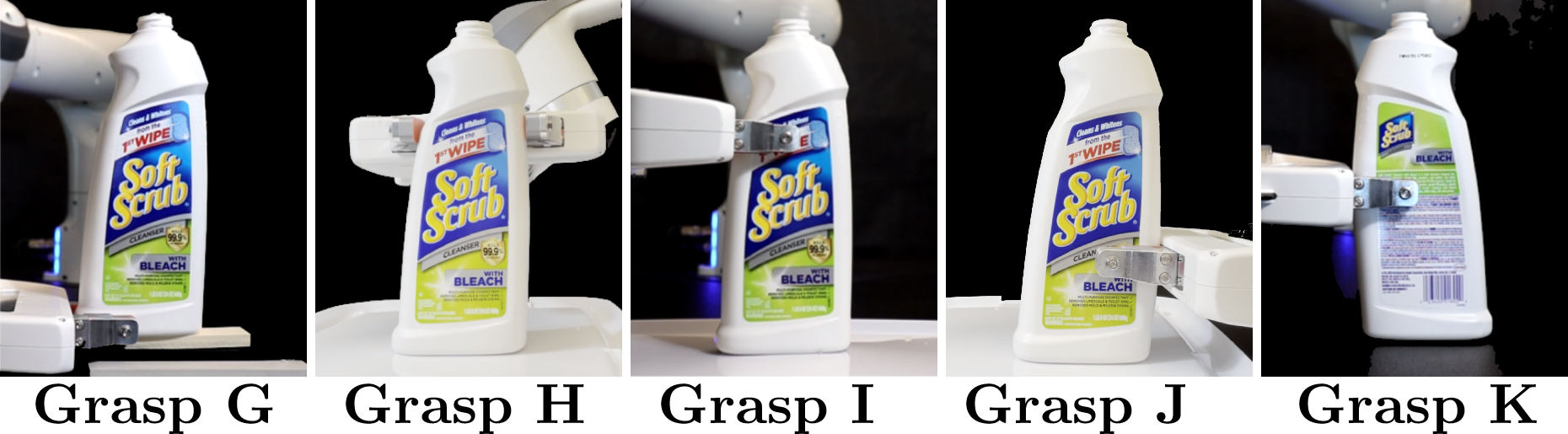}
\caption{Five tested grasps on a real bleach bottle. Grasps are also repeated in simulation.}
\label{fig:bleach_grasps}
\end{figure}

\begin{figure}[h]
\centering
\includegraphics[scale=0.45, trim={0cm 0cm 0cm 0cm},clip]{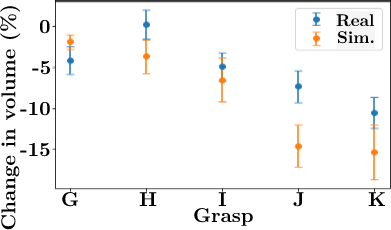}
\caption{Percent volume change pre- and post-grasp for the real and simulated bleach bottles under grasps G to K.}
\label{fig:bleach_volume}
\end{figure}

\noindent\textbf{Bleach bottle}. We also evaluate five grasps on a real bleach bottle (Fig.~\ref{fig:bleach_grasps}), which are ordered from G to K in ascending order of deformation imparted on the simulated version of the bleach bottle. Since deformation fields on the bottle are not readily accessible in the real world, we instead measure the volume of the bottle. After applying a fixed grasp force, the bottle is filled with rice, and the weight is recorded and divided by density. The resulting volume change of the real-world grasps is similar to simulated results, except that simulation incorrectly predicts that grasp H would impart more deformation than grasp G (Fig.~\ref{fig:bleach_volume}). Cutting open the bottle reveals that material at the neck of the bottle is thicker (1~mm) than at the bottom (0.85~mm), whereas a uniform wall thickness is assumed in simulation. Thus, local stiffness higher on the bottle (including the contact region under grasp H) may be underestimated. This discrepancy in wall thickness also explains why the simulated grasps J and K predicted significantly more change in volume than in real life, as the geometric stiffness was underestimated within simulation.  

\begin{figure}
\centering
         \includegraphics[width=\linewidth, trim={0cm 0cm 0cm 0cm},clip]{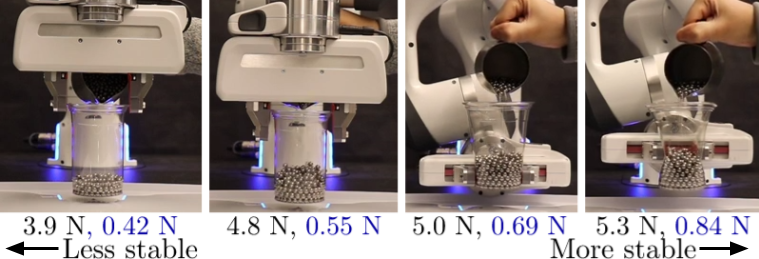}
         \caption{Under four grasps on a plastic cup, the maximum weight withstood before loss of contact is annotated for both the real world (black) and in simulation (blue). 
}
\label{fig:cup_stability}
\end{figure}

\noindent\textbf{Plastic cup}. We test 4 grasps on a plastic cup for stability. In simulation, we run the linear acceleration experiment in the upward direction; in the real world, the cup is gradually filled with metal balls until contact is lost. Since the real Franka has no precise force controller, the actual gripper forces on the cup are unknown (unlike the prismatic tofu, the cup has complex geometry that makes force hard to analytically estimate from position inputs). Thus, while the ordering of grasps with respect to stability is consistent between real world and simulation, the weight values at failure are not (Fig.~\ref{fig:cup_stability}).

\section{Conclusions and Future Applications}
We propose and motivate performance metrics to describe deformable grasping outcomes as well as grasp features to serve as potential predictors of metrics. We then measure these quantities by conducting a battery of grasp simulations on 3D deformable objects and release our dataset of 6800 grasps and $1.1M$ measurements for further study, along with software that executes our experiments on arbitrary objects and material parameters. Finally, our simulated results are also shown to have good correspondence with real-world grasp outcomes. We envision DefGraspSim to be a useful research tool for the community working towards grasping deformable objects, with direct applications to:

\begin{itemize}
    \item Learning representations of new high-dimensional features and metrics (e.g., for object and contact geometry, field quantities, etc.) for memory-efficient grasp planning
    \item Customizing grasping experiments to create task-oriented planners (e.g., to minimize food deformation)
    \item Performing rigorous, direct comparisons between simulation and reality on custom deformables of interest (e.g., on organs for robotic surgery)
    \item Generating training data for real-world system identification (e.g., tactile probing on unknown materials) 
    \item Generating data for neural network-based grasp simulation for real-time grasp planning
    \item Improving grasp planning robustness to uncertainty in object material properties (e.g., via domain randomization)
\end{itemize}

\vfill\null

\ifCLASSOPTIONcaptionsoff
  \newpage
\fi

\bibliographystyle{IEEEtran}

\bibliography{references}

\end{document}